\documentclass{article}
\usepackage{spconf,amsmath,epsfig,caption}
\usepackage[utf8x]{inputenc}
\usepackage{url}
\usepackage{color}

\usepackage{graphicx,epstopdf}
\DeclareGraphicsExtensions{.ps}
\usepackage[colorinlistoftodos]{todonotes}

\usepackage[ruled]{algorithm2e}
\usepackage{scalerel}
\newcommand\vfrac[2]{\ThisStyle{%
  \setbox0=\hbox{$\SavedStyle#1#2$}%
  \setbox2=\hbox{$\SavedStyle X$}%
  \ifdim\ht0>\ht2\setlength{\ht0}{\ht2}\fi%
  #1\mathord{\stretchto{\raisebox{2.3\LMpt}{$\SavedStyle/$}}{\ht0}}#2}}

\title{ Three dimensional Deep Learning approach for remote sensing image classification}
\name{Amina Ben Hamida* **, Alexandre Benoit*, Patrick Lambert*, Chokri Ben Amar**}
\address{*Univ. Savoie Mont Blanc, LISTIC,
~~\\ F-74000 Annecy, France
~~\\(amina.ben-hamida,alexandre.benoit, patrick.lambert)@univ-smb.fr
~~\\
**
REGIM, Ecole Nationale d’Ingénieurs de Sfax
~~\\Route de Soukra, B. P. W, 3038 Sfax, Tunisie
~~\\chokri.benamar@ieee.org}
\begin{document}
%
\maketitle
\begin{abstract}
Recently, a variety of approaches has been enriching the field of Remote Sensing (RS) image processing and analysis. Unfortunately, existing methods remain limited faced to the rich spatio-spectral content of today's large datasets. It would seem intriguing to resort to Deep Learning (DL) based approaches at this stage with regards to their ability to offer accurate semantic interpretation of the data. However, the specificity introduced by the coexistence of spectral and spatial content in the RS datasets widens the scope of  the challenges presented to adapt DL methods to these contexts. Therefore, the aim of this paper is firstly to explore the performance of DL architectures for the RS hyperspectral dataset classification and secondly to introduce a new three-dimensional DL approach that enables a joint spectral and spatial information process. A set of three-dimensional schemes is proposed and evaluated. Experimental results based on well known
hyperspectral datasets demonstrate that the proposed method is able to achieve a better classification rate than state of the art methods with lower computational costs.
\end{abstract}
\begin{keywords}
Remote Sensing, Hyperspectral, Deep Learning, pixel-based, Classification
\end{keywords}
\section{Introduction}
Today, Remote Sensing (RS) plays a fundamental role in providing a rich source of information for a variety of applications. It is now a major means for advanced and numerous purposes such as the long-term climate studies, population evolution analysis and sometimes even the precocious prevention of calamities. In fact, RS has opened doors for not only a deeper understanding of the Earth itself but also for delicate investigations into its population and environmental behaviors. In reality, such advances are mainly boosted by the collaboration of serious industrial and academic research. The impressive breakthroughs witnessed on a technical level, acquisition tools as well as new open data models, have dramatically impacted RS data products. This fast progress has mainly led to the creation of not only overwhelming quantities of datasets but also of very rich spatial and spectral content. Faced with harder and more complex images, the renovation of the classically used approaches has been needed. In fact, the basic philosophy behind the first RS classification methods relied on the so called "shallow structures". Different techniques have been used including artificial neural networks \cite{Kav}, classification trees \cite{Mc} and support vector machine \cite{SVMchaussot}. The Bag of Visual Words (BoVW) was introduced as a baseline for recent RS image classification \cite{Bids} since it enables a better understanding of the data content and the inter-pixel dependencies. Although these tools were -until recently- highly ranked in the classification field, they are now incapable of coping with the abundance of today's image content. As a response to this lack of efficient methods, serious efforts have been made and several approaches have been set up, such as the graphic based one detailed in \cite{Lefevre} or some handcrafted feature tools that can effectively describe the spatial and spectral content of the images \cite{11IGARS}. Despite high performance, these approaches remain limited because of their lack of a generic aspect, adaptation to different contexts and an impassable need for expert knowledge in the parameter set up phase.  Therefore, there is an urgent need for more convenient analysis methods and approaches that allow hierarchical comprehension of the data and thorough learning of its content. Certainly, when dealing with tricky learning tasks, it is currently almost impossible not to acknowledge the achievements of Deep Learning (DL). In fact, since the impressive comeback of neural networks in 2006, the machine learning community has become very popular, mainly thanks to the emergence of DL based approaches and their remarkable performances. Early deep methods started in simpler ways with digit classification to recently become a winning tool for complex image classification tasks in the 2012 Large Scale Visual Recognition Challenge (ILSVRC2012). Over the past several years, DL has been growing as one of the most efficient techniques for a wide range of applications and fields and has managed to overcome different challenges when dealing with Big Data issues. However, it likewise brings more precision and accuracy into smaller scale applications that mainly focus on today's wealth of spatial and spectral content. Therefore, one of the main focuses is now pointed at the ability of DL approaches to solve RS data classification problems. Currently, an important share of RS research is devoted to investigating techniques that enable effective interpretation, analysis and extraction of relevant knowledge. 

 In this paper, a general overview of the DL evolution phases and the currently existing methods is presented. The main challenges that are disrupting its progress are also presented along with a special focus on DL techniques used for RS image classification. Finally, a new deep network structure is proposed and examined for an RS case study on a small well-known hyperspectral dataset which then draws the basic guidelines of how to deal with the RS data blast. Models and trained weights are then made available at \url{https://github.com/AminaBh/3D_deepLearning_for_hyperspectral_images}.

\section{Deep Learning over the years}
Over the years, a lot of research has been dedicated to machine learning and artificial
intelligence. Obviously, talking about neural networks is not a new subject as the field has been around since the 1950s \cite{Perceptron}. Starting from the mid-90s, machine
learning has gone through several transition periods paving the way for the impressive
comeback of Neural Networks. However, the proposed techniques relied so much on human involvement in the process for system tuning and data annotation along with high computational power requirements that neural nets where surpassed by other less constrained methods such as support vector machine based approaches. Consequently, without sufficient resources, neural network-based approaches went through a "winter period" where few advances appeared in state of the art methods. Thanks to Geoffrey E. Hinton and his team, the world migrated from shallow structures to deep architectures with the introduction of Deep Belief Networks (DBNs) in 2006, which have revolutionized both the academic and industrial worlds. This progress has been greatly assisted by  technical evolution on different levels: the world has witnessed the birth of richer annotated databases along with powerful computational tools such as GPUs. Since then, much literature has been focused on DL methods. The main tool behind the success of DL is the introduction of more processing layers which induces more representational levels and therefore, ensures progressive dissociation of the concepts contained in the data. Consequently, not only does DL enhance the learning process, these approaches have also managed to overcome the classical ones by getting rid of the previously used engineered features. Over the years, a rich repository  has been established to encompass numerous deep architectures. These methods can be categorized into 3 main classes regarding their architectures, aims and techniques that are detailed in the following subsections.

\subsection{Deep networks for unsupervised learning: Generative approaches}
The absence of target labels or knowledge information during the learning phase limits the application to a feature identification process. Most of today's available data are unlabeled datasets which raises the question: to what extent can we learn meaningful representations? In this case, the lower-level abstractions are more tightly related to simple features and higher level abstractions are dedicated to high semantic level concepts and objects. Different approaches belong to this specific category namely autoencoders \cite{encod}. Basically, They can perform image hierarchical learning through two types of modules: a first set of data encoding layers followed by a decoding set of layers that tries to reconstruct back the input. Another proposed architecture for unsupervised learning is the Restricted Boltzmann Machine \cite{RBM}. As first introduced, this approach relied on connected neuron-like units that make stochastic decisions about whether to be on or off. Therefore, it can be seen as a neural network model \cite{BM}. Although the main concept behind it seems very tempting for image processing applications, its execution was difficult and time-consuming. As a remedy for this problem, restrictions were added to the network topology forbidding connections between the variables within the same layer and leading to the one layer restricted Boltzmann machines (RBMs, \cite{BM34}) as well as its deeper version the Deep Boltzmann
machine (DBM). Deep Belief Networks (DBNs) where then proposed in \cite{DBN} and rely on a clever combination of RBMs along with a classifier.
\subsection{Deep networks for supervised learning: Discriminative approaches}
Target labels are expected to help learning and data classification, whether they are present in a direct or indirect form. This category of approaches is intended to accomplish pattern classification tasks, often by characterizing the posterior distributions of classes presented in the data. They are also called discriminating deep networks.
Deep Stacking Networks (DSN), Convolutional Neural Networks (CNNs)\cite{Lecun} and Recurrent Neural Networks \cite{RNN} are the main architectures used for supervised tasks.
As first introduced, the main idea behind the DSN design derives
from the concept of arranging series of simple classification modules, as proposed and
explored in \cite{28} and \cite{44}. At an early stage, classifiers are set to then be stacked on top of each other ensuring the learning of complex concepts. The DSN architecture was originally presented in \cite{Deng}. Although different varieties of DSNs were created in order to diminish the computational cost of the process, these architectures remain very expensive, which challenges their users and deflects interest toward other sets of approaches. One of the main alternatives is the Convolutional Neural Network (CNN), as first presented in \cite{Lecun}. These Convolutional Layers (Conv) can be viewed as a series of trainable filters that slide all over the input's dimensions (width, height and even depth). Since these layers share many weights but significantly less than classical fully connected neuron based networks, the stacking of the layers on top of each other allows a gradual increase in the data representation semantic level without exploding the computational cost. Usually, a series of fully connected layers followed by a classifier are inserted at the end of each phase. This basically finalizes the representation learning process by modeling the target concepts from a composition of their already high semantic level input features. The CNNs benefit from the bonus of introducing the sub-sampling property which guarantees a decrease in feature map resolutions thus reducing the computing costs while enforcing robustness against translations. In fact, discriminative approaches generally follow a similar philosophy that aims at first representing input images as high resolution low level features representations, starting by oriented contours, that are gradually sub-sampled and composed into more complex and more numerous patterns while going threw the network architecture.
Depending on the task, a final layer may be used to format output to the required
type. Luckily, all these basic tools have allowed the creation of a rich benchmark of robust designs for different CNN networks. These architectures have been found highly effective and been commonly used in computer vision and image recognition \cite{55}, \cite{57}.

\subsection{Hybrid deep networks}
The term hybrid for this third category refers to the deep architecture that either comprises or makes use of both generative and discriminative model components. These architectures often operate in a multistage learning process, where the generators are trained using a specific strategy. The recent Generative Adversarial Networks \cite{Adv} highlights the interest of simultaneously training the two components where a generator continuously improves and tries to fool a discriminator which continuously tries to differentiate real and fake generated data. Recently, it was shown in \cite{ORB} that deep hybrid architectures, or multi-level models that integrate discriminative and generative learning objectives, offer a strong viable alternative to multi-stage learners.
\subsection{ The evolution of different CNNs}
As one of the most successful deep architectures, CNNs have progressively found their way into today's applications. By less by 20 years, we moved from the 5 layered leNet5 architecture \cite{Lecun} dedicated to digits recognition to the more advanced Residual Network \cite{Resnet} variants that can include hundreds of layers that are now able to recognize thousands visual concepts. Actually the main architecture that effectively familiarized the world with CNNs
is the 8-layer AlexNet. As detailed in \cite{krizhevsky2012}, this technique has the advantage of being deeper and more expressive than the other approaches by stacking a series of Convolution layers. The AlexNet architecture was a clear winner at the ILSVRC challenge of 2012 and since then, this challenge
has been systematically won by CNNs every year, always improving performance levels by improving depth, width, processing path strategies while reducing the number of parameters \cite{ZF, Google}.
Then, as the world has evolved towards more sequence-dependent applications (video, text, etc.), Recurrent Neural Networks(RNN) whose output depends on the input and the previous iteration states complete CNN architectures. The Long Short-term Memory (LSTM) cells \cite{botvinick2006} is a flexible example of such family that enables long and short range interactions by the use of trainable state gates. Such tools enable impressive results on various application use cases such as image captioning \cite{captioning14} and video semantic segmentation \cite{videosemanticsegm16}.

\section{Deep Learning for Remote Sensing image Classification}

The content of satellite images with high resolution in both space and frequencies is
remarkably complex, providing details of objects like houses,
trees, or even cars on the parking lots. In order to be able to
fully describe the content of these images, a deep hierarchical representation
is highly recommended. The lowest level is represented by primitive vectors
describing the color, texture and shape. At a higher level, simple objects like roads, forests or lakes are described by unique combinations
of primitive vectors. However, individually considered, these objects
can not describe the scene or grasp the overall meaning of the image as they give different interpretations according to their neighboring, as well as their spatial and spectral positions. Therefore, in order to extract meaningful information from the image, the spatial interactions to the next level of the representation hierarchy must be taken into account. These models lead to the discovery of semantic rules that define the final level of abstraction: high level semantic classes like residential districts, commercial areas or ports are of high semantic level. This extremely rich repository of information has created the need for DL as a key solution.

\subsection{State of the Art DL for RS}

It is often possible to resort to ideas that come from the multimedia field to proceed with RS datasets. Current multimedia inspired DL models have managed to provide a baseline for the use of DL in RS key applications. Recent challenges focus on semantic segmentation and ensure high accuracy rates in the case of a traditional 3 spectral bands task \cite{PASCALVOC}. However, the creation of more complex RS data has catalyzed more research into better understanding of data with rich spectral content such as hyperspectral and multispectral images. As detailed in \cite{IGARS}, first trials only relied on the spectral information
presented in the data itself. For instance, the approaches as presented in \cite{ILV}, \cite{79IGARS} and \cite{80IGARS} suffered a lack of spatial information and therefore probably disregarded a very important element of the image content. The same problem was seen when only processing the spatial content of the data as detailed in \cite{Fauvel}, \cite{Huang2015}, \cite{Zhao2016155} and \cite{Zhang201675}. Therefore, more complex yet effective solutions have been presented in different forms and models to take into account both spectral and spatial components guaranteeing a maximum profit from the insights and information restrained in the images. Early solutions resorted to a marginal processing of spectral and spatial information as presented in \cite{SAE1}. In this case, the spectral information is processed apart from the spatial component which is extracted later to be joined for feature extraction based on deep architectures like stacked autoencoders (SAE). Neural network classifiers are then implemented in the final layer. Auto encoders are also the basic concept behind the model introduced in \cite{SAE2}, presenting a spatio-spectral framework that merges spectral information from adjacent pixels to add spatial information to the processed pixel. Hidden layers are then inserted in order to learn the spectral features and a supervised learning is ensured by an output softmax layer. Incorporating both spatial and spectral information improves the classification performances as mentioned in all methods above. However, the use of SAE or more generally using large layers of fully connected neurons explode the number of parameters to train and demand a large number of training samples. In the case of datasets with few annotations, training systems with a large number of parameters is not tractable or lead to over-fitted and sub-optimal solutions. 
In the specific case of hyperspectral image analysis, DL methods recently opened wide doors into taking huge leaps. More specifically, 3D (i.e. one spectral dimension plus two spatial ones) CNNs were introduced as a solution to obtain an accurate and computationally efficient architecture. As presented in \cite{makantasis}, a 3D like approach starts the process with a randomized PCA applied on the spectral dimension of the image. Therefore, the inputs for the first convolutional (Conv) layer (C1) are 3D patches of size $ s \times s \times Cr$ where s is the width and height of the spatial patch while
Cr is the number of the retrained principle spectral components. A second Conv layer is applied to the output of C1. Finally, a C2 element vector is produced and fed as input to a Multi-Layer Perceptron (MLP) classifier. However, with this approach, each retrained spectral dimension is processed independently with standard 2D convolutional filters. Another approach is developed in \cite{Viktor}, where a 3D convolution is literally deployed by the first layer followed by two 1D Convs and ending with two Fully Connected Layers (FC). This approach is close to the one presented in \cite{Hindawi} but adds spatial information. However, it still introduces a large amount of parameters (60000) that need to be trained with only 1800 samples. 
The main concern in most of the cases presented above is then how to deal with the high number of parameters to be trained with few samples while improving image analysis.
Recently, architectures significantly reduce this ratio such as \cite{densenet} that maximizes the parameter reuse. 

\subsection{CNNs for RS image Classification}
In view of the wealth of recent RS image content, it is crucial to find deep
architectures that maintain the balance between efficiently processing huge amounts of data and not exploding the computational costs while also providing high accuracy. The employment of deep classification techniques for the RS field would seem to be a promising path of applications. However, further investigation reveals different challenges that must be overcome to reach accurate low cost data interpretation. The main challenge is to efficiently adapt deep architecture to take into account not only the spatial dimension of hyperspectral images but also their rich spectral content. Out of all the current DL networks, CNNs are one of the best available tools for machine vision. These models have helped DL become one of today’s hottest topics. Thanks to the variety of layers one CNN can encompass, these networks provide an efficient tool for data comprehension and representation. The fact that they can be fitted to different applications and are relatively low cost architectures for tremendous tasks make them one of the most extensively used DL approaches. In this paper use cases, CNNs are a primordial choice that can be fitted to RS classification tasks. Basically, these architectures must perform well without over-fitting or under training the system. However, the evolution towards effective CNNs in such cases has been diminished by the following challenges.
\begin{itemize}
\item {\textbf{High dimensional data}}
\newline
When dealing with high dimensional data, DL
approaches become computationally-expensive. These high costs are mainly due to the
slow learning process that is needed to learn the data abstractions and establish an
effective representation from low levels to the highest semantic interpretations. In fact, a high-dimensional data source contributes greatly to the volume of the raw data, as well as complicating learning from this data. The most effective solution so far proposed is the use of CNNs since the neurons in the hidden layer units do not need to be connected to all of the nodes in the previous layer, but to a more localized receptive field. Moreover, the resolution of the image data is also reduced when moving toward higher
layers in the network as depicted in \cite{Bengio91}, \cite{Bengio2003} thanks to pooling layers. However, this problem remains very challenging and leads to more complex and harder issues. 
\item{\textbf{Large heavy models}}
\newline
DL models have so far accomplished remarkable results relying on deep and wide  models. Therefore, large numbers of parameters are required to learn complicated features and representations from the data itself as explained in \cite{widez}. These heavy models  are hard to train, costly to fit and complicated to establish. Moreover, such heavy models are greedy in terms of labeled data. This requirement is hard to establish since the field is suffering from a serious lack of rich hyperspectral and multi-spectral annotated data. 

\item{\textbf{Architecture Optimization}}
\newline
The key point in favor of using DL today is
its ability to cope with a wealth of applications. However, this results in hardening and complicating the tasks of establishing deep models that are inexpensive and effective in processing data. Obviously, regarding
the variety of fields that today's community is involved in, one can notice an urgent need to optimize deep architectures in terms of computational costs,
accurate results and required training information. Recently various strategies have been proposed to optimize pre-trained architectures in terms of inference speed and memory footprint reduction by layer factorization and weights pruning \cite{Han2015a,Han2016a, Parashar2017}. However, such post training optimization strategies cannot allow to disregard architecture design optimization prior training since large architectures cannot be reliably trained on small datasets. Therefore, a lot of wise choices must be taken at early stages: selecting one specific type of deep network, choosing whether to fine-tune a pre-trained architecture or starting from scratch.
\end{itemize}

The CNN structure introduces an accurate solution for most of the previously presented challenges. However, more recent studies that tend to enhance the use of deeper architectures have also been established. Residual modules as presented in Residual Networks \cite{Resnet} have made it possible to design extremely deep networks with more than a thousand layers. More recently, Dense Networks were presented in \cite{densenet} that emphasize the interest of denser connections across layers. The main purpose of such methods is generally to address complex problems with high variability targets at multiple scales. Recent works tried to adapt these concepts to hyperspectral data use cases as detailed in \cite{Igarss}. The aim of this paper is to prove the efficiency of non-complex architectures.

\section{ 3D Deep Learning architecture for RS image Semantic segmentation}

As a solution for the challenges presented in the previous sections, we introduce a new three dimensional based architecture that is dedicated to hyperspectral images and tackles most of the DL for RS aspects of difficulty.

\subsection{General overview of the architecture}

\begin{figure}[htp]
  \centering
  \includegraphics[width=8.5cm, height=4cm]{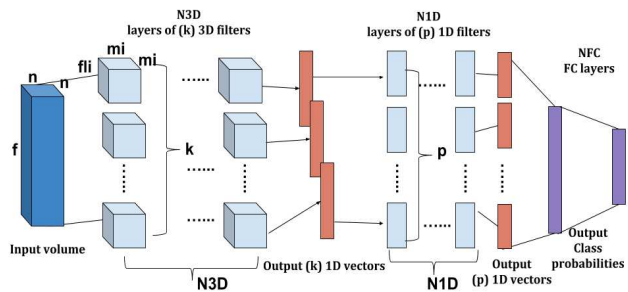}
  
  \caption{Overview of the proposed 3D Deep architecture.}
  \label{fig:3d}
\end{figure}

A joint spatio-spectral model is needed to examine both spectral and spatial information in hyperspectral data. The advantage of such a framework is that both the components are merged and joint in a non separable way from the early stages of the process. This solution makes maximum use of the information presented in the data and radically lowers costs. This paper proposes to use a new 3D CNN architecture that, unlike the previously mentioned approaches, simultaneously processes the spatial and spectral components with real 3D convolutions giving better investments of the few samples available with fewer trainable parameters. This proposal decomposes the problem as the processing of a series of volumetric representations of the image. Therefore, each pixel is associated to an\textit{ $n \times n$} spatial neighborhood and a number of \textit{f} spectral bands. As a result, each pixel is treated as a 
\textit{$n \times n \times f$} volume. The main concept behind this architecture is to combine the traditional CNN network with a twist of applying 3D convolution operations instead of using 1D convolution operators that only inspect the spectral content of the data. An overview of the 3D architecture is presented in Figure \ref{fig:3d}.

Different blocks of CNN layers are stacked on top of each other in order to ensure deep efficient representations of the image. Firstly, a 3D convolution based set of layers is introduced in order to cope with the three dimensional input voxels. Each and every one of these layers encompass a number of volumetric kernels that simultaneously execute convolutions on the width, height and depth axis of the input. Such 3D convolutions stack is followed by a set of \textit{$1 \times1$}  convolution (1D) layers that discards the spatial neighborhood and a series of Fully Connected layers. Basically, the proposed architecture considers 3D voxels as input data and first generates 3D feature maps that are gradually reduced into 1D feature vectors all along the layers. This procedure is ensured by the choice of specific configurations of the convolution filter strides and paddings following equation \ref{eq:sizeout}, where the stride is the distance between two consecutive positions of the kernel expressed with a number of pixels. The padding is used to manage boundary effects and is basically employed as a number of zeros concatenated at the beginning and at the end of an axis. Using padding enables to make the convolution output the same size as the input while no padding reduces the output data shape. 
 \begin{equation}
 \label{eq:sizeout}
 SizeOut =\left( \displaystyle\frac{SizeIn-Kernel Size + 2\times pad}{stride}\right)+1
 \end{equation}
As previously said, the input is fed to the network as a 3D volume (voxel) of size \textit{$n \times n \times f$}. The first phase consists of using a series of \textit{N3D} 3D convolutional layers. Each layer $\textbf{\textit{i}}$ is characterized by a number of  \textit{ki} filters. The kernels of the filters are of size \textit{($mi\times mi\times fli$)} where \textit{mi$<=$n}
and \textit{fli$<=$f}. In this case, convolution layers are deployed for two purposes. First, they are introduced as conventional spatio-spectral convolution layers with a stride equal to 1. Then, they play the role of pooling layers thanks to the choice of larger strides to down sample the data as suggested in \cite{simplicity}. The duality between Conv and Pool layers in a sequential way progressively learns and reduces the data components' dimensions. The \textit{$fli>mi$} rule along with the removal of padding on the spatial dimension for some layers make the transition towards the creation of a first 1D output vector. This output is then fed to a series of  \textit{N1D} 1D Conv layers that each encompasses \textit{pi} filters. In the end, the network introduces a set of \textit{NFC} Fully Connected (FC) layers that ends with a Softmax classifier where the softmax activation of the i\textsuperscript{th} output unit is detailed in (2). Since the final layer's size is chosen to be equal to the input number of targeted classes (nclass), this FC guarantees a probabilistic representation for the different classes. 
  \begin{equation}
    P(x_i) = \displaystyle\frac{e^{x_i}}{\displaystyle\sum_{c}^{nclass} e^{x_c}}
  \end{equation}
  where $x_i$ denotes the 1D output vector of the final Conv layer.
  
  \begin{figure}
  \centering
  \includegraphics[height=6.5cm]{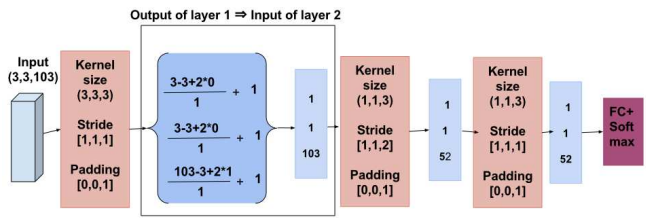}
  \vspace{-1cm}
  \caption{Example illustrating the evolution of feature shapes (sizeOut) of each layer (output size is obtained according to Eq.  \ref{eq:sizeout}).}
  \label{fig:size}
\end{figure}
An example of a 3D Conv layer network is illustrated in Figure \ref{fig:size} where, $f=103$, $n=3$, $fli=3$ and $mli$ can take 3 or 1 as values. Padding is used on the spectral scale only. The strides are alternated between one and two in order to create a pooling effect after every convolutional operation.
\begin{algorithm}
    \SetKwInOut{Input}{Input}
    \SetKwInOut{Output}{Output}
    \Input{$S$: number of samples per batch ;\\ $epochIT\_{train}$: number of iterations required to parse all the train dataset with $S$ samples per batch (one epoch);\\ $epochIT\_{test}$: number of iterations required to parse all the test dataset with $T$ samples per batch (one epoch);\\$MaxEpoch$: Maximum number of epochs to train the network;\\ $X_{train}$: Input training voxels;\\ $X_{test}$: Input testing voxels}
    \Output{mAccuracy: the average accuracy on the test dataset;\\
    All trained weights}
\For{\texttt{epoch in range $MaxEpoch$}}{
 \textbf{Testing:} evaluate accuracy for each voxel from $X_{test}$\\
 \For{\texttt{it in range $epochIT\_{train}$}}
        {  
        \hspace{0.15cm}1) Randomly sample $S$ voxels from $X_{train}$ not already been considered in the current epoch;
        \newline{2) Forward propagation of the $S$ input voxels through the network in order to calculate the average loss; }
        \newline{3) Backward propagation to update network weights with respect to the average loss value;}
        }
 \textbf{Testing:} evaluate accuracy for each voxel from $X_{test}$\ 
\newline
\For{\texttt{it in range $epochIT\_{test}$}}
        { \hspace{0.1cm} 1) Sample $T$ voxels from $X_{test}$ not already been considered in the current epoch;
        \newline{2) Run a forward propagation on the batch samples.}
        \newline{3) Retrieve and cumulate the obtained accuracy values}}
compute $mAccuracy$, the average accuracy values on the entire test dataset obtained in the current epoch.
}
\label{algo:1}
    \caption{Classification of Hyperspectral images at the pixel level. Each pixel is processed w.r.t. its neighborhood only as a voxel of shape $n\times n\times f$.}
    
\end{algorithm}

So, as a summary and as detailed in Algorithm 1, the proposed process starts by dividing the hyperspectral image into two different sets of pixels: a training set of pixels and a testing one. In fact, each pixel is taken into account as a 3D $n \times n \times f$ voxel in the paper's context. A training period is considered to parse one time all the pixels of the training dataset (one epoch) to learn the network weights. Follows a test period that evaluates the performance level reached on the whole test dataset. Those two periods are applied $MaxEpoch$ times to get the final result. In this way, the evolution of the network performance is monitored all along the training.
\begin{figure*}[htp]
  \centering
  \includegraphics[width=\textwidth,height=8cm]{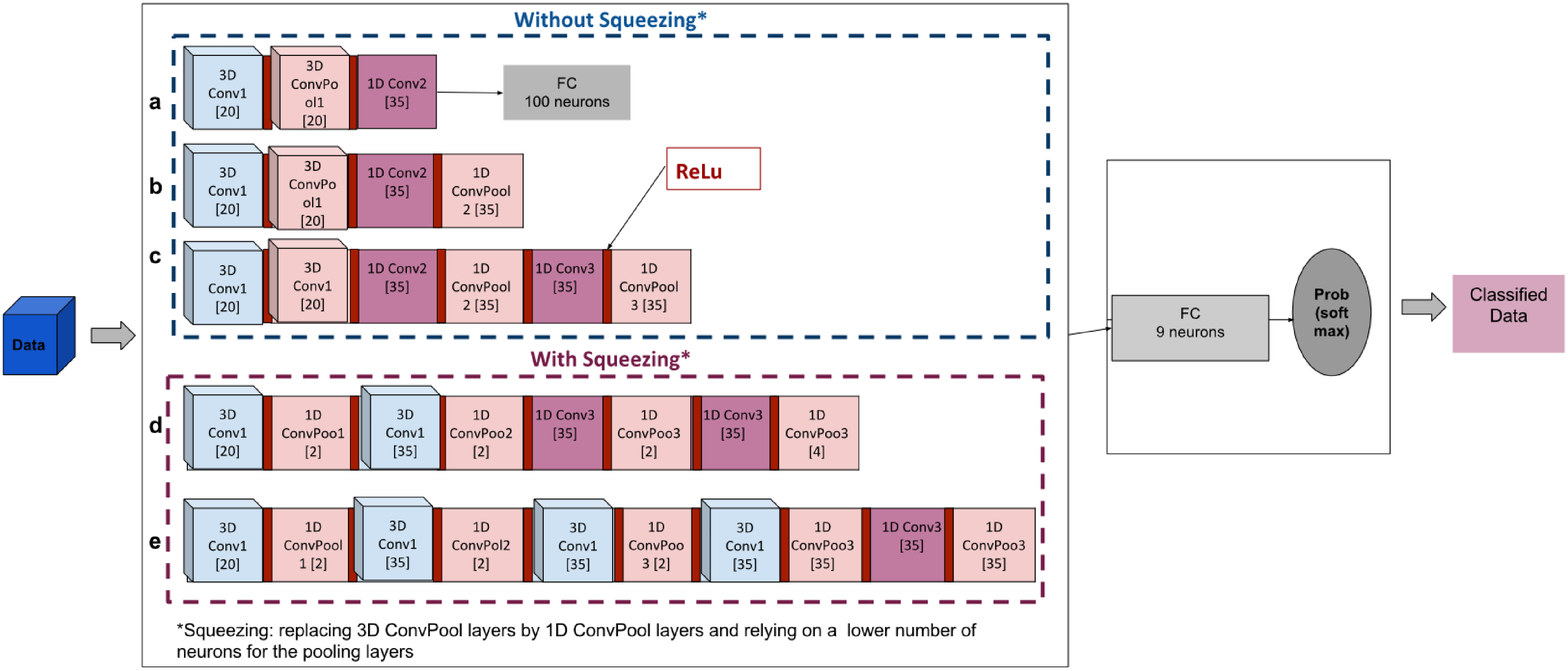}
  
  \caption{Overview of our 3D architectures.}
  \label{fig:Overview}
\end{figure*}
\subsection{Main common parameters}
When establishing a DL architecture, the most crucial phase is to make wise choices for the different parameters to be set. Although different models can be obtained from one basic DL architecture, one can easily notice that there are common parameters that can be fulfilled from the early stages of the network creation. 
\begin{itemize}
\item {\textbf{Solver method}} plays a major role in improving loss during the learning process and contributes in both the forward and backward propagation phases and therefore ensures the update of the network's parameters. Different solver methods have been introduced so far such as the Stochastic Gradient Descent (type: ``SGD"), Adam (type: ``Adam"), Nesterov’s Accelerated Gradient (type: ``Nesterov") and RMSprop (type: ``RMSProp") as detailed in \cite{SGD} and \cite{adam} . The Stochastic Gradient Descent ``SGD" with momentum set to 0.9 was selected for this use case. Several tests with different solver types were the reason behind this choice which appears to be a simple and robust method.

\item {\textbf{Weights regularization method}} is basically used to introduce lateral regularization to the network. The main types of normalization methods are L1 and L2 weights regularization which both provide the choice to either normalize within the same channel or even normalize across channels. L1 normalization proved to be more efficient in the paper's context. One recent regularization technique is the dropout method \cite{dropout} which randomly disables some inputs thus reducing neuron dependencies. This also complements L1 and L2 regularization by preventing the network from over fitting. Therefore, we have combined the use of both L1 and a 0.5 probability dropout on the Fully Connected layer only.
\item{\textbf{Non-linearity}} as detailed in the previous section, most of the attention today is dedicated to the use of ReLU non-linearities since they enable faster training convergence. Recent non-linearities such as ELU \cite{ELU} have also been experimented but did not lead to improved results.
\item{\textbf{Weight Initialization}} is a crucial pre-processing phase for DL network training since it sets the state of different parameters at the starting point. Different methods can be used including the recent \textit{Xavier} and MSRA \cite{weight} methods. The MSRA was chosen since it is adapted to the ReLU non-linearities that we used.
\item{\textbf{The learning rate}} coordinates to what extent each step of the process influences the weight updates. We consider an initial learning rate of 0.001 to explore rapidly the search space and find a good local minimum. Then, each $\vfrac{MaxEpoch}{3}$ iteration, learning rate is divided by 10 in order to converge to a lower local loss value and then increase accuracy (MaxEpoch is the number of epochs needed to train the network).
\item{\textbf{The batch size}} instead of optimizing the network from a single sample at a time, which can lead to sub-optimal solutions, averaging errors over a set of samples was proved to be more efficient. Therefore, batch size influences the loss convergence efficacy and in turn the system's update phase, which makes it a very critical parameter and basically data dependent. Values ranging from 10 to 1 were tested on the small considered dataset and the optimal choice is equal to 3.
\item{\textbf{Bias or batch normalization:}} batch normalization  has recently been proposed to normalize neuron activation across layers and replace neuron bias variables. Despite being efficient, batch normalization requires consistent and stable batch statistics in the training and test database. This is often helped by the use of a large batch size. This constraint is difficult to fulfill on tiny databases so each time we restricted our experiments to neuron layers equipped with bias parameters initialized to 0 at the beginning of the training.
\end{itemize}

\section{Experiments and analysis}
In this section, the experiments conducted on RS images using DL architectures are presented and compared to other state of the art approaches.
\subsection{The datasets}
\begin{figure}
  \centering
  \includegraphics[width=9.5cm,height=4.5cm]{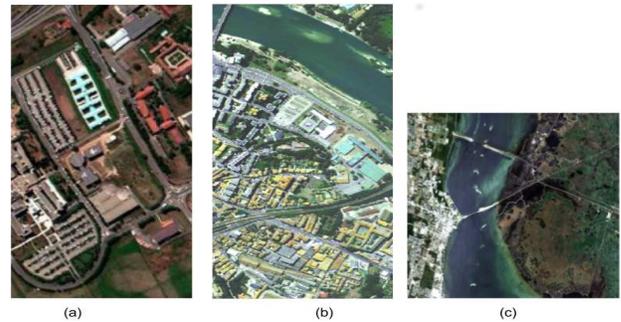}
  
  \caption{Captured Images: (a) Pavia U,(b) Pavia C and (c) Kennedy Space Center.}
  \label{fig:datasets}
\end{figure}

The results are obtained from experiments applied on the University of Pavia, the Pavia Center and the Kennedy Space Center datasets, all collected by the AVIRIS sensor and shown in RGB colors in Figure \ref{fig:datasets}. The first employed data was a capture of an area over Pavia University, northern Italy, with a spatial resolution of 1.3 m. The image comprises\textit{ $610 \times 340$} pixels with 103 bands. Next, the Pavia Center is a 102-band dataset that presents one image of size \textit{ $1096 \times 1096$} pixels and of 1.3 m geometric resolution. Finally, the Kennedy Space Center dataset was acquired over the Kennedy Space Center (KSC), Florida, USA, on March 23, 1996. This image has 224 bands from 400 to
2500 nm and the spatial resolution is 18 m. After removing water
absorption and low signal-to-noise (SNR) bands, it has
\textit{ $512 \times 453$} pixels  with 176 bands. The first two datasets include some challenging scenes among the 9 classes which are respectively Water, Trees, Asphalt, Self-Blocking Bricks, Bitumen, Tiles, Shadows, Meadows and Bare Soil for Pavia University and Asphalt, Meadows, Gravel, Trees, Painted Metal Sheets, Bare Soil, Bitumen, Self-Blocking Bricks and Shadows for the Pavia Center dataset. The KSC dataset include 13 classes which are respectively Scrub, Willow swamp, Cabbage palm hammock, Cabbage palm/oak hammock, Slash pine, Oak/broadleaf hammock, Hardwood swamp, Graminoid marsh, Spartina marsh, Cattail marsh, Salt marsh, Mud flats and Water.
\vspace{-0.48cm}
\subsection{Different 3D architectures}
Extensive sets of experiments were conducted. The main approaches that best summarize the performance level of the 3D architectures are reported below. The most performing models and trained weights are then made available at \url{https://github.com/AminaBh/3D_deepLearning_for_hyperspectral_images}. The strategy is to start from a simple state of the art architecture and then gradually extend it by understanding its bottlenecks. Figure \ref{fig:Overview} presents an overall view of the considered deep networks where the Conv layers with a stride equal to 1 are referred to as \textbf{Conv} and those with a stride equal to 2 as \textbf{ConvPool}. The number of filters per layer is introduced as [number of filters]. So, in Figure \ref{fig:Overview}, each layer is presented as follows: ``LayerName[number of filters]". The a, b, c, d and e schemes respectively represents the 3, 4, 6, 8 and 10 layer-networks as detailed below.
\begin{itemize}
\item {\textbf{3 layers VS 4 layers: a VS b}}
\newline
As first inspired from the \cite{Hindawi} network, the presented architecture created a 3-layer 3D network that gathers two 3D layers, one single 1D layer along with two FC layers and a Softmax. However, the important number of neurons included in the first FC layer increases the number of parameters without really improving the accuracy rates. This layer is basically dedicated to increase the capacity of the final classifier while taking care of the spatial arrangement in the feature maps. However, the increased number of parameters conflicts with the low number of training samples. Therefore, in a new 4-layer network, only one FC, with the number of neurons equal to the number of targeted classes, is left and a new Conv layer with a stride equal to two is added to play the role of a pooling layer. The network is then able to keep up with the same performance level with a lower training cost.
\item {\textbf{6 layers network: c}}
\newline
The noticeable drop in the number of trained parameters in the 4-layer network provided an interesting opportunity to develop deeper models with more Conv layers. Therefore, one 6-layer 3D architecture was created relying on two 3D Conv layers followed by 4 1D layers, where the duality Conv/ConvPool is sequentially applied to the network. Finally, one FC with a Softmax was kept. The number of filters for each layer was set to 35 except for the first Conv that only gathered 20 filters. In fact, a wide spectrum of filter numbers with values ranging from 5 to 50 filters per layer was tested. The optimal combination is then presented in this paper. This choice goes with the standard state of the art tendency that shows that fewer filters are required at the beginning of the architecture.

\item {\textbf{Squeezing the net: towards deeper architectures: d and e}}
\newline
Since different deep models can guarantee the same accuracy level, the choice among them is then based on the cost and number of parameters each network can take. Therefore, as inspired from the SqueezeNet presented in \cite{squez}, the trials to compress the model led to the creation of lighter models. Thanks to the use of smaller numbers of filters along with $1 \times 1 \times 3$ filters in the pooling phase, the network can reach the same accuracy levels with a smaller number of parameters. Indeed, \textit{$1 \times 1$} Conv layers with a number of neurons lower than the one in the previous standard convolutional layer (2 neurons for ConvPool VS 35 neurons for Conv) allows a significant reduction of the number of parameters. As a result, deeper architectures were created for better learning. 
\item{ \textbf{8 and 10-layer networks: d and e}}
A deeper light architecture has been created with fewer parameters. A 8-layer 3D network was created first introducing the duality between 3 3D Conv layers and 3 1D ConvPool layers, followed by 2 1D layers and a single FC with a softmax. The same goes for a 10-layer network with the advantage of adding one more sequence of 3D layers along with pooling ones.
\item{ \textbf{Networks width vs depth}} Talking about the depth of the deep networks is a very rich debate that generates a lot of questions. However, it has been recently proved \cite{widez} that one of the keys for better performances is to find the right balance between the network's depth and width. In other words, fixing the number of layers per network is a very crucial step that has a huge impact on the efficiency of the architecture. But, knowing that the network has a variant number of filters per layer catalyzes the concerns about which number to choose. Therefore, estimating the appropriate number of filters for each layer (the width of the network) according to its depth is an important decision to be taken in order to harmonize the cost and accuracy of a deep network. In this paper, the first layer is characterized with 20 filters while the rest of the layers have 35 filters, except for the pooling layers that have 2, 4 or 8 filters according to their position in the network as discussed in the above ``Squeezing the net" section. 
\end{itemize}
\subsection{Experiments and Results}

Experiments and tests were executed on the 3, 4, 6, 8 and 10 Conv layered architectures as detailed in the previous section. For the sake of an accurate comparison with state of the art methods, two main data splitting strategies were used. When proceeding with the 3 and 4-layer networks, only 200 randomly chosen pixels per class were used for training (almost 4\%) while the rest of the data pixels were kept for the testing phase. Then, in the case of deeper architectures, 5\% of the images were deployed for training using the same class balance strategy. Each model is trained and evaluated 3 times using different non overlapping train and test random splits. Accuracy levels are averaged to report a synthetic performance measure. We measured a redundant 0.2\% precision error and do not report this value within the tables to lighten the presentation. All the tests were executed on a 4-core intel i7-6600U laptop CPU with no GPU included. The presented results were obtained using the caffe library \cite{Caffe}.
\begin{itemize}
\item \textbf{3 layers VS 4 layers : a VS b}

First, as detailed in \cite{Bids}, the introduction of the 3D architecture is inspired from \cite{Hindawi}. Technically, two main differences distinguish the proposed architecture from the one detailed in \cite{Hindawi}. First, trainable convolution layers with strides greater than one are used instead of the classically used max pooling filters. Then, the 100 neuron FC introduced by \cite{Hindawi} is replaced by a 50 neuron FC. However, the accuracy difference cannot only be explained by these factors. In fact, as detailed in Table \ref{Tab:3layers}, if the number of neurons in the FC layer is divided by 2 (50 neurons vs 100 in \cite{Hindawi}), the parameter number decreases by almost 18\% while the performance level remains stable. This shows that the initial system has actually too many degrees of freedom which misleads the image representation and limits the system efficiency. Although these primitive results do not enhance the level of the accuracy rate, they provide us with hints towards establishing better models. The first observation to be taken into account is that the choice of the spatial neighborhood is very crucial and data dependent. Here, the \textit{$5 \times 5$} input spatial neighborhood seems to be the optimum choice for the Pavia University dataset while the \textit{$3 \times 3$} one performs better in the cases of the Pavia Center and Kennedy Space Center datasets. However, the \textit{$7 \times 7$} spatial neighborhood seems to be very extended compared to the spatial components of the datasets which makes it a low performer. Furthermore, the decrease in the number of neurons in the first FC layer results in a decrease in the number of the overall trained parameters without influencing the accuracy rates.

\begin{table*}
\centering
\captionof{table}{Accuracy level of the 3-layer CNN models (a) using 4.4\% of the data for training.}\label{Tab:3layers}
\begin{tabular}{ |p{1cm}||p{1cm}|p{1cm}|p{1cm}|p{1cm}|p{1cm}|p{1cm}|p{1cm}|p{1cm}|p{1.2cm}|p{1cm}|p{1cm}|}

 \hline
network& Conv Nb & 3D Conv Nb  & 1D Conv Nb  & Spatial neighborhood &Spectral depth & FC Nb  & Fc neurons Nb & Pavia university& Pavia Center  & paramet-er Nb  & Training time(s) \\
 \hline

 a&3 & 0 & 3 & \textit{$1\times1$}&3&2&[50,9]&   75.9\%  &90.4\%  &51669 &1386  \\
 \hline
a&3 & 2 & 1 & \textit{$3\times3$}&3&2&[50,9] &79.3\% & \textbf{96.7\%} &53189&432 \\
 \hline
a&3 & 2 & 1 & \textit{$5\times5$}&3&2&[50,9]& \textbf{89.2\%}& 94.5\%  & 56669&10257 \\
 \hline
a&3 & 2 & 1 & \textit{$7\times7$}&3&2&[50,9]&   85.9\%  & 96.2\% &61949&24465 \\
 \hline
  \hline
\cite{Hindawi}&3 & 0 & 3 & \textit{$1\times1$}&3&2&[100,9]&   92.5\%  & None &None&None \\
 \hline
\end{tabular}
 
\end{table*}
\begin{table*}
\centering
\captionof{table}{Accuracy level of the 4-layer CNN models (b) using 4.4\% of the data for training.}\label{Tab:4layers}
\begin{tabular}{ |p{1cm}||p{1cm}|p{1cm}|p{1cm}|p{1cm}|p{1cm}|p{1cm}|p{1cm}|p{1cm}|p{1.2cm}|p{1cm}|p{1cm}|}

 \hline
network& Conv Nb & 3D Conv Nb  & 1D Conv Nb  & Spatial neighborhood &Spectral depth & FC Nb  & Fc neurons Nb & Pavia university& Pavia Center  & paramet-er Nb  & Training time(s) \\
 \hline

 b&4 & 0 & 4 & \textit{$1\times1$}&3&1&[9]&   75.0\%  &93.4\%  &17759 &1203  \\
 \hline
b&4 & 2 & 2 & \textit{$3\times3$}&3&1&[9] &84.0\% & \textbf{97.1\%} &27524&360 \\
 \hline
b&4 & 2 & 2 & \textit{$5\times5$}&3&1&[9]& \textbf{93.8\%}& 96.4\%  & 28749&9000 \\
 \hline
  \hline
\cite{Hindawi}&3 & 0 & 3 & \textit{$1\times1$}&3&2&[100,9]&   92.5\%  & None &None&None \\
 \hline
\end{tabular}

\end{table*}
\vspace{0.5cm}
As detailed in Table \ref{Tab:4layers}, the removal of the first FC layer along with the introduction of the spatio-spectral concept in the network enables better results compared to \cite{Hindawi}. In fact, in this case the data representation is more relevant when going from 3D voxels to 1D vectors followed by a single nc-neuron FC layer (nc = number of the dataset classes) combined with a Softmax. Not only does this model benefits from better accuracy rates, it also witnesses an important decrease on the computational cost level. Going down from more than 60,000 parameters trained in \cite{Hindawi} to 28,749 parameters, the proposed 4-layer network ensures both a better learning of the data content and a lower training cost process. Here again, the results prove that the choice of the spatial neighborhood is a \textit{$5 \times 5$} for the Pavia University and a \textit{ $3 \times 3$} for the Pavia Center.
The reliance on the 3D Conv layers for combined spatio-spectral classification of the data is therefore a basic key for better results when compared to \cite{Hindawi} that only resort to the spectral signature of each pixel in the classification process.

\item \textbf{6-layer network: c}

Establishing a 6-layer network was inspired by the important decrease in the number of parameters in the 3 to 4-layer transition case. The introduction of a sequence of Conv and Pooling duets in the network gather both the benefits of going deeper and involving fewer parameters. In other words, more Conv layers ensure higher semantic level representation of the data while the Pooling ones guarantee a dimension reduction of the representation. This way, the dimension of the vectors at the entry of the FC layer is remarkably reduced thus significantly reducing the number of parameters. As shown in Table \ref{Tab:6layers}, an important decrease in the number of parameters (from a tenth of the previous 60,000) is witnessed, along with an increase in the accuracy rate. These tests also prove that the choice of the spatial neighborhood is highly dependent on the data content. The same model can outperform the \cite{Lefevre} results in the case of the Pavia Center dataset with a \textit{$3 \times 3$} neighborhood while it does not reach the state of art approach results in the case of Pavia University even when using \textit{$5 \times 5$} neighborhood.

\begin{table*}
\centering
\captionof{table}{Accuracy level of the 6-layer CNN models (c) using 5\% of the data for training.}\label{Tab:6layers}
\begin{tabular}{ |p{1cm}||p{1cm}|p{1cm}|p{1cm}|p{1cm}|p{1cm}|p{1cm}|p{1cm}|p{1cm}|p{1.2cm}|p{1cm}|p{1cm}|}

 \hline
network& Conv Nb & 3D Conv Nb  & 1D Conv Nb  & Spatial neighborhood &Spectral depth & FC Nb  & Fc neurons Nb & Pavia university& Pavia Center  & paramet-er Nb  & Training time(s) \\
 \hline

 c&6 & 0 & 6 & \textit{$1\times1$}&3&1&[9]&   86.5\%  &94.1\%  &4754 &1500  \\
 \hline
c&6 & 2 & 4 & \textit{$3\times3$}&3&1&[9] &90.9\% & \textbf{98.5\%} &6074&600\\
 \hline
c&6 & 2 & 4 & \textit{$5\times5$}&3&1&[9]& \textbf{94.6\%}& 97.08\%  & 6074&12000 \\
 \hline
  \hline
\cite{Lefevre}&- & - & - & -&-&-&-&   \textbf{98.1\%}  & 97.0\% &None&None \\
 \hline
\end{tabular}

\end{table*}

\begin{table*}
\centering
\captionof{table}{Accuracy level of the 8-layer CNN models (d) using 5\% of the data for training.}\label{Tab:8layers}
\begin{tabular}{ |p{1cm}||p{1cm}|p{1cm}|p{1cm}|p{1cm}|p{1cm}|p{1cm}|p{1cm}|p{1cm}|p{1.2cm}|p{1cm}|p{1cm}|}

 \hline
network& Conv Nb & 3D Conv Nb  & 1D Conv Nb  & Spatial neighborhood &Spectral depth & FC Nb  & Fc neurons Nb & Pavia university& Pavia Center  & paramet-er Nb  & Training time(s) \\
 \hline

 d&8 & 0 & 8 & \textit{$1\times1$}&3&1&[9]&   90.3\%  &94.7\%   &2161&940  \\
 \hline
d&8 & 2 & 6 & \textit{$3\times3$}&3&1&[9] &92.9\% & \textbf{98.9\%} &3681&286\\
 \hline
d&8 & 2 & 6 & \textit{$5\times5$}&3&1&[9]& \textbf{97.2\%}& 98.1\%  & 6862&7605 \\
 \hline
  \hline
\cite{Lefevre}&- & - & - & -&-&-&-&   \textbf{98.1\%}  & 97.0\% &None&None \\
 \hline
\end{tabular}

\end{table*}
\begin{table*}
\centering
\captionof{table}{Accuracy level on Kennedy Space Center dataset using 5\% of the data for training.}\label{Tab:KSC}
\begin{tabular}{ |p{1cm}||p{1cm}|p{1cm}|p{1cm}|p{1cm}|p{1cm}|p{1cm}|p{1cm}|p{1.2cm}|p{1cm}|p{1cm}|}

 \hline
network& Conv Nb & 3D Conv Nb  & 1D Conv Nb  & Spatial neighborhood &Spectral depth & FC Nb  &  Fc neurons Nb& Accuracy&  paramet-er Nb  & Training time(s) \\
 \hline

 b&4 & 2 & 2 & \textit{$3\times3$}&3&1&[13]&   71.01\%    &18698&105  \\
 \hline
c&6 & 2 & 4 & \textit{$3\times3$}&3&1&[13] &77.5\% & 1175&98\\
 \hline
d&8 & 2 & 6 & \textit{$5\times5$}&3&1&[13]& \textbf{84.2\%}& 2251&145 \\
 \hline

\end{tabular}

\end{table*}
\begin{figure*}[htp]
  \centering
  \includegraphics[width=16cm,height=7cm]{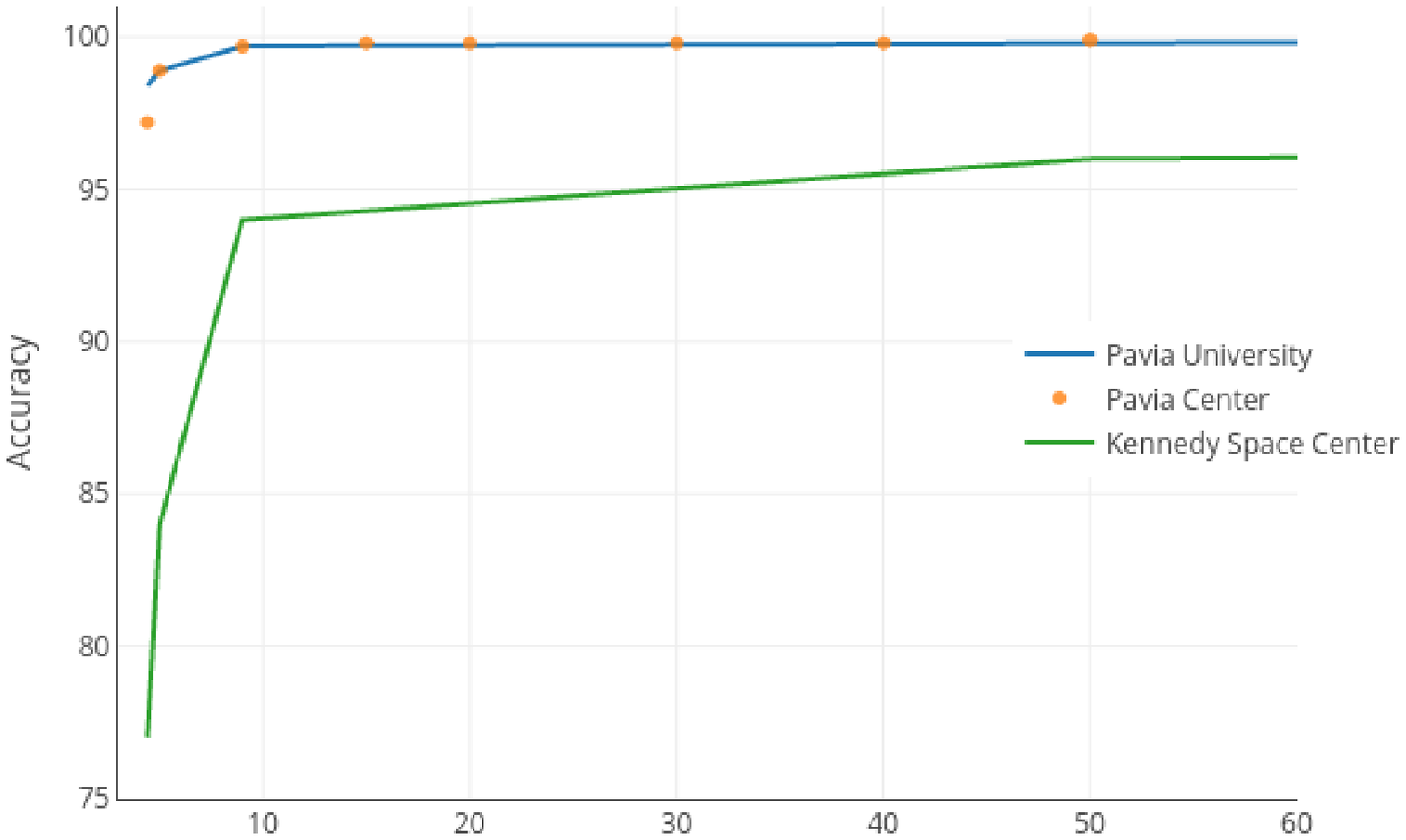}
  \caption{Accuracy (\%) VS  amount of training data (\%) using the 8-layer CNN model (network d).}
  \label{fig:Accuracy}
\end{figure*}
\textbf{\item Squeezing the net, 8 or 10-layer networks: d or e.}

According to the results presented in the previous three tables, the main key behind a successfully performing deep network is to create a balance between a deep yet light architecture. When examining the models, it can be seen that the number of layers in the network (the depth) and the number of neurons per layer (the width) manipulate the most important share of the network performance. Therefore, in Tables \ref{Tab:SqueezePaviaU} and \ref{Tab:SqueezePaviC} the tests were executed for different combinations of a number of neurons per layer, a number of 3D layers per architecture and the overall number of layers in the network. Although a wider range of values was tested for each parameter presented below, only the most effective ones are mentioned at this stage.

\captionof{table}{Squeezing the net on Pavia University: (c), (d) and (e) using 5\% of the data for training.}\label{Tab:SqueezePaviaU}

\begin{tabular}{ |p{0.6cm}|p{0.6cm}|p{1.5cm}|p{1cm}|p{1.2cm}|p{1cm}|}

 \hline
  Conv Nb& 3D Conv Nb & Pooling neuron Nb& paramet-er  Nb& Accuracy& Training time (s)  \\
 \hline
 6 & 2   & [2,2,35]&11123&94.6\%& 9000
 \\
 \hline
6 &   2  & [2,2,8]&4680&86\%&8700   \\
 \hline
6 &2 & [2,2,4] & 3931&92.3\%&8514
 \\
 \hline
 8 & 2 & [2,2,35,35]& 27470&96,3\%&10000\\
 \hline
  8 & 3 & [2,2,35,35]& 20993& 96.8\%&10834\\
 \hline
 8 & 3 & [2,2,2,35]&  16340 & 94.1\%&9324\\
 \hline
  8 & 3 & [2,2,2,4]& 6862 & \textbf{97.2\%}&\textbf{7605}\\
 \hline
  10 & 3 & [2,2,35,35]& 40740 & 94.3\%&11474\\
 \hline

\end{tabular}

\captionof{table}{Squeezing the net on Pavia Center:  (c), (d) and (e) using 5\% of the data for training.}\label{Tab:SqueezePaviC}
\begin{tabular}{ |p{0.6cm}||p{0.6cm}|p{2.1cm}||p{1.8cm}|p{1.1cm}|}

 \hline
Conv Nb &  3D Conv Nb& Pooling neuron Nb & parameter Nb& Accuracy \\
 \hline
 6 & 2   & [2,2,35]&11123&96.1\%
 \\
 \hline
6 &2 & [2,2,4] & 3931&92.8\%
 \\
 \hline
 8 & 2 & [2,2,35,35]& 27470&98,3\%\\
 \hline
  8 & 3 & [2,2,2,4]& 6862 & \textbf{98.9\%}\\
 \hline
\end{tabular}

\vspace{0.5cm}
As detailed in the tables above, 
The decrease in the number of neurons per pooling layer enables lighter models that maintain the same accuracy rate ranges. Basically, the decrease in the width of the network provides more opportunities to create deeper models. Therefore, the 6-layer network performs less than the 8 and 10-layer ones. However, the 8-layer architecture seems to be the best choice especially when relying on three 3D layers, along with a small number of neurons in the pooling layers. Not only does it reduce the number of parameters but it also enhances the accuracy rate. 

The best derived combination in our case is an 8-layer network with 3 3D blocks. The number of filters are equal to 20 for the first Conv layer and increased to 35 for the following Conv layers. However, the numbers of neurons for each Pooling layer are respectively equal to 2,2,2 and 4. Here, we proceed with an input voxel of size [5,5,103]. Table \ref{Tab:8layers} highlights the best performing architectures in the case of 8-layer networks.
\begin{figure*}[htp]
  \centering
  \includegraphics[width=\textwidth,height=6.5cm]{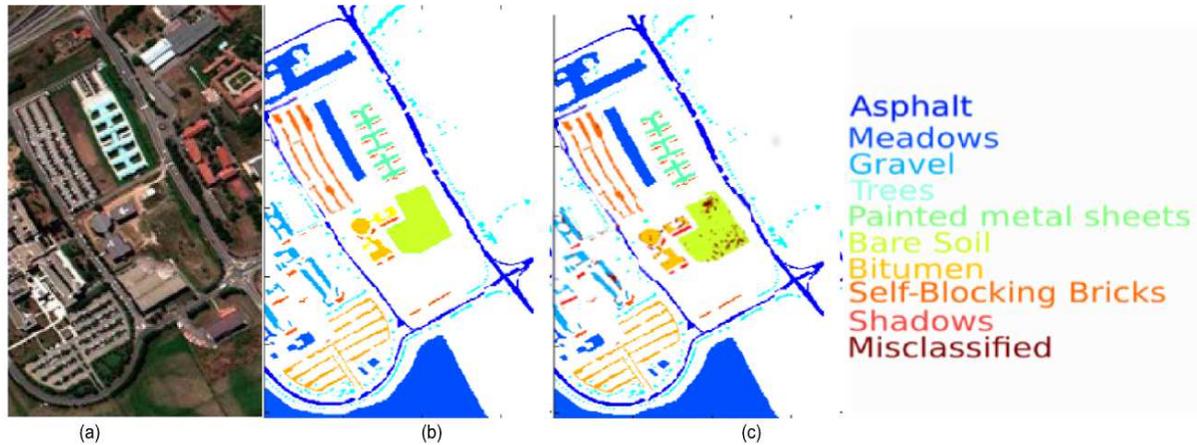}
  \caption{The Pavia University dataset: (a)  Image, (b) Ground Truth, (c) Classification output 6-layer network.}
  \label{fig:segResultsExample}
\end{figure*}

\subsection{Architecture Selection}

The introduction of different architectures in the previous sections enables different performance levels. At this stage, the choice of an optimal network is made. As previously detailed, many factors interfere with the selection process of a best performing network. In fact, the progressive evolution toward a deeper, yet lighter network, draws the guideline toward an easier choice. 

The first presented architectures (a and b), managed to establish a solid baseline for the creation of deeper 6, 8 and 10-layer Networks. Since, the c, d and e models enable high accuracy rates, the computational cost plays then a major role in the choice making process. As detailed in tables 1, 2, 3, 4, 5, 6 and 7, the 8 layer network enable the highest accuracy rates for all of the 3 datasets. Besides, it ensures low computational costs with lower training parameter numbers. Therefore, for what comes next, the 8 layered-architecture will be used. As seen in Figure \ref{fig:segResultsExample}, the display of the output classification map doesn't give much information about the system's precision since the accuracy rates overpass the 90\% in most of the 6 and 8 layers cases of study. Therefore, the confusion Matrices are drawn and presented in Figure \ref{fig:confusionMat} in order to demonstrate the performance level of the networks. In this figure, the first two datasets represents classes ranging  from 0 to 8 and the KSC confusion matrix represents classes ranging from 0 to 12 according to their enumeration order used in section 5.1. Although  these matrices show highly accurate classification rates, they also demonstrate the confusions made by the trained network as shown in Figure \ref{fig:confusionMat}. In the case of the Pavia University dataset, the Bitumen pixels are mixed up with the Shadow pixels with almost 1/10 mistaken pixels among all the classified ones. The lowest accuracy rate (almost 60\%) is witnessed in the case of the Kennedy space Center Image, where the Cabbage palm/oak hammock class is confused with 4 other classes: The mistaken pixels are classified as 20\% oak/broadleaf hammock, 10\% Cabbage palm hammock and less than 10\% divided between slash pine and water. The system behavior toward such classes is basically driven by the high spectral and spatial similarity between the two class characteristics.

\captionof{table}{Processing time required to reach 95\% of the final accuracy on the Pavia University dataset.}
\begin{tabular}{ |p{2cm}||p{1cm}|p{1.3cm}|p{1.1cm}|p{1.1cm}|}
 \hline
 Architecture& Iterat-ions Nb & Processing Time (s)& Accuracy (early)& final Accuracy (100\%) \\
 \hline
  8 layers $1\times1$ & 23733 & 313&85.8\%&90.3\%
 \\
 \hline
8 layers $3\times3$  & 23500 & 95&88.3\%&92.9\%
 \\
 \hline
 \textbf{8 layers $5\times5$  }& \textbf{23233} & \textbf{2535} &\textbf{92.4\%} & \textbf{97.2\%}\\
 \hline
  6 layers $5\times5$  & 23233 & 40000&89.8\%&94.6\%\\
  \hline
   4 layers $5\times5$  & 23233 & 3000&89.1\%&93.8\%\\
 \hline
\end{tabular}

  \begin{figure}
  \centering
  \includegraphics[height=6.5cm]{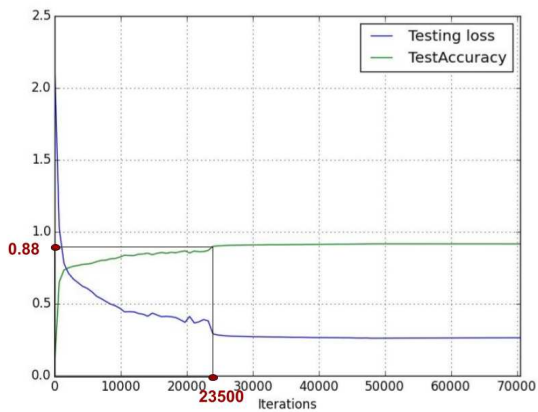}
  \vspace{-1cm}
  \caption{Processing time required to reach 95\% of the final accuracy on the Pavia U dataset using the 8-layer $3 \times 3$ network.}
  \label{fig:convergence}
\end{figure}
\vspace{0.5cm}
In order to get better insights on the choice of a given architecture, taking a look at the accuracy and loss values along the training phase is critical. Actually, these indicators can validate the choice of hyper-parameters, such as the learning rate, but it also shows the training stability and how an architecture reaches a stable performance level. Plotting those curves for several architectures can however be confusing, added to the fact that they are not perfectly reproducible due to random weights initialization. Then, for readability, we report the monitored values obtained at a specific iteration close to the maximum accuracy value but before the steady state. It somehow represents a form of early stopping in the training process where system performances are estimated. More specifically, we select the iteration index that reports 95\% of the final accuracy value. Note that averaging those values along multiple experiments enhances related values confidence. Figure \ref{fig:convergence} shows an example of such monitoring report on a single experiment. One can also observe a step at iteration 22000 for both the loss and accuracy curves. This actually corresponds to the iteration where a decrease in the learning rate occurs. This change suddenly improves accuracy since a good local minimum has been found previously and then convergence to a lower loss value is made possible. In the proposed networks setup, this step happens few iterations before the proposed architectures report 95\% of their final accuracy value. Finally table 8 makes the synthesis of the obtained results on the most performing architectures. As a conclusion, the 8 layer-architectures enable high accuracy rates within a reasonable period of time in the paper context. One can also notice the impressive convergence speed of the 8 layer architecture relying on $3\times3$ pixels neighborhood. Such architecture enables access to high accuracy values at very low cost and is then the best compromise to obtain rapidly a good classifier in environment with low energy and processing time limitations.

\begin{figure*}[h!]
  \centering
  \includegraphics[width=\textwidth,height=6cm]{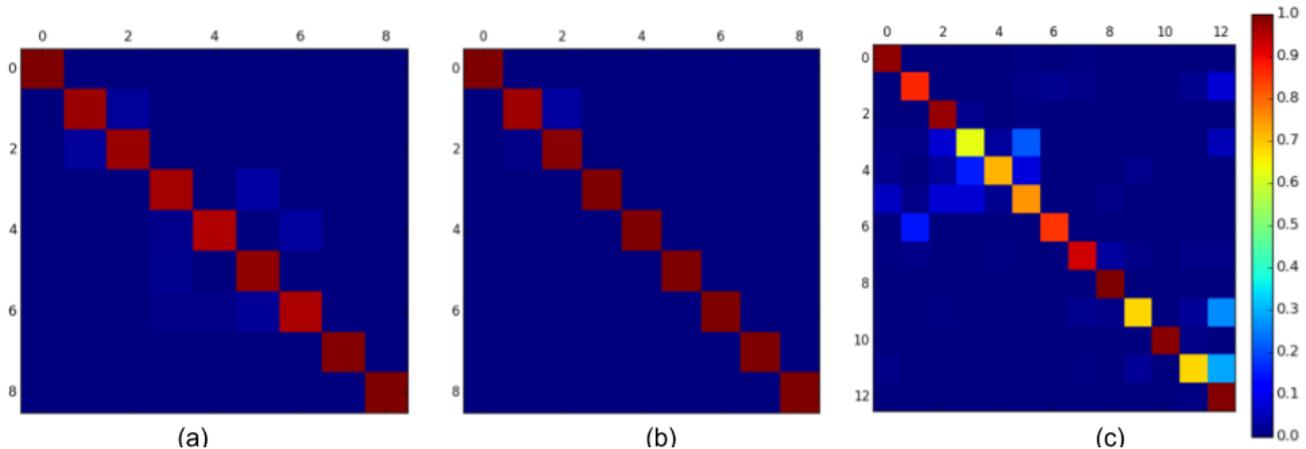}
  \caption{Confusion Matrix using the 8 layered-network (a)  Pavia U (b) Pavia C (c) KSC.}
  \label{fig:confusionMat}
\end{figure*}

\subsection{Computational cost and training data requirements}
Obviously, the developed deep architecture over-performs the existing methods in the case of the Pavia Center dataset. Unlike different approaches such as \cite{newpaper}, \cite{review1} and \cite{review2}, the 3D proposed architectures enable from scratch-training \textbf{with no prior data preprocessing}. Not only does it simplify the processing task, the 3D architecture enables the same performance levels as the previously mentioned approaches. Besides, the time taken is a main factor when evaluating DL architectures. Here, the classification process takes almost 3 hours at the most with a single Intel i7-6600U laptop CPU (no GPU used) in the case of heavy architectures along with relatively large dataset images. However, it only takes less than 2 minutes in the case of light networks with few training samples. Another key factor that normally influences the performances of these classification processes is the amount of data involved for the training phase. For example, when using only 5\% of the image pixels for training, the 3D architecture can't overpass the 98\% accuracy rate in the case of the Pavia University dataset (even when using the most performing architecture). However, when dedicating 10\% of the data for training as experimented in \cite{newpaper}, the accuracy rates in that case are more than 99.4\%. As detailed in  \ref{fig:Accuracy}, the 3D architecture is capable of reaching a \textbf{99\% accuracy rate while only using 9\% of the image pixel to train the network.} The state of the art method detailed in \cite{SAE1} also deploys 9\% of the dataset for training. However, this approach under-performs the proposed 3D architecture with a 98\% accuracy level for about 20,000 trained parameters against our 99\% accuracy level for less than 7,000 trained parameters. 

\subsection{Model transferability}
The previous sections demonstrate the possibilities of establishing a new light Deep Neural Network that takes into account both the spectral and the spatial raw data. It has been proven so far that this 
architecture performs well in the case of hyper-spectral data. However, regarding the lack of richly annotated hyper-spectral images, this paper and more generally state of the art methods only examine and review training and testing in the same context. First, the training is executed using a subset of a single image that is specific to a given context. Then, testing and inference is based on the remaining pixels of the same image, i.e we generalize on the same context. However,  many questions can arise from this strategy, since the deep learning models are highly exposed to the possibility of over-fitting, so, did the model over-fit on each specific data context ? Can the learned features be transfered from one dataset to the other ?\\
Therefore, in what follows, the learned feature transferability between different contexts is proposed, trained using one specific image and evaluated based on a second image. Since the target classes are the same, transfer learning between Pavia University and Pavia Center is proposed. However, the number of spectral bands differs (103 bands vs 102 bands) so that the output dimensions of the convolution layers will differ from one dataset to the other. In this context, it is necessary to resort to architecture fine tuning: given a neural network architecture trained on a given dataset, all its architecture components (except the last fully connected layers) are kept and their weights are made constant. Finally, the fully connected layers are replaced with new ones whose number of connections (weights) are compatible with the new dataset and the shape of the convolution layer output. A rapid training of the new network component is performed on the new dataset on its own training set and performances are evaluated on the test set.
\begin{table}
\centering
\captionof{table}{Accuracy level for Fine-tuning using 5\% of the data for training.}\label{Tab:finetune}
\begin{tabular}{ |p{1.5cm}||p{1.5cm}|p{1.2cm}|p{1.2cm}|p{1.2cm}|}

 \hline
Train data& Test data& Network& spatial size&Accuracy \\
 \hline

 Pavia U&Pavia C & d & $3 \times 3$ &98.4\%  \\
 \hline
Pavia C&Pavia U & d & $3\times 3$& 90.4\% \\
 \hline

\end{tabular}

\end{table}
As detailed in Table \ref{Tab:finetune}, the use of a pre-trained 8-layer model in the case of the Pavia University and Pavia Center datasets provide an accurate pixel wise classification of the data. In fact, the Deep Neural Networks are able to maintain nearly the same precision level when fine-tuned and trained from scratch. (98.4\% VS 98.9\% and 90.4\% VS 92.9\%). Basically, the pre-trained architectures proposed in this paper demonstrate a strong ability to generalize to other context images. As in the case of the Pavia University and Pavia Center datasets, the re-use of pre-trained networks saves the huge effort required to re-create a specific architecture to be trained from scratch from each and every use case.
\end{itemize}

\section{Conclusion}
The processing of hyperspectral data in general is a very delicate procedure that demands the effective use of both spatial and spectral components. The benefit of the 3D architecture introduced in this paper is to not only accurately classify the hyperspectral data but also to establish deep comprehension of the images at low cost. One of the most valuable consequence is the ability to efficiently optimize deep networks on small sized annotated datasets which then also reduce the cost of the data. The main concern now is to investigate ways to innovate and enhance the created models in order to process with larger, heavier datasets. As a remedy for such a challenge, both Residual and Dense Networks enable the fusion of different representation levels. Therefore, they would seem like an appealing solution to enhance the existing CNN architectures. Furthermore, Hyperspectral data calibration is still an open issue that currently confines its use to limited areas. An interesting path would therefore to create architectures that are able to cope with this issue.

\bibliographystyle{IEEEbib}
\bibliography{ref_trans}

\end{document}